\PassOptionsToPackage{table}{xcolor}

\documentclass{article}

\usepackage{microtype}
\usepackage{graphicx}
\usepackage{subfigure}
\usepackage{booktabs} 

\usepackage{hyperref}
\usepackage{seqsplit}

\usepackage{enumitem}
\usepackage[accepted]{icml2025}

\usepackage{amsmath}
\usepackage{amssymb}
\usepackage{mathtools}
\usepackage{amsthm}
\usepackage{multirow}

\usepackage{tabularx}

\usepackage[capitalize,noabbrev]{cleveref}

\usepackage[most]{tcolorbox}  
\usepackage{xcolor}           
\definecolor{myblue}{RGB}{0,102,204}  

\usepackage[table]{xcolor}    
\usepackage{booktabs}          
\usepackage{multirow}          
\usepackage{arydshln}          
\usepackage{graphicx}          
\usepackage{listings}
\usepackage{microtype}

\usepackage{pifont}
\usepackage{hhline}

\usepackage{tabu}
\usepackage{fontawesome5}

\usepackage{nccmath}
\usepackage{algorithm}
\usepackage{caption}
\usepackage[algo2e, ruled, vlined]{algorithm2e}

\SetKwComment{Comment}{$\triangleright$\ }{}
\SetKwProg{Init}{Initialize}{}{}

\usepackage{algorithmic}
\renewcommand{\algorithmicrequire}{\textbf{Input:}}
\renewcommand{\algorithmicensure}{\textbf{Output:}}
\renewcommand{\algorithmiccomment}[1]{\bgroup\hfill//~#1\egroup}

\usepackage[export]{adjustbox}
\usepackage{caption}

\usepackage{wrapfig}

\newcommand{\lparagraph}[1]{\textbf{#1}~}

\usepackage{xspace}
\newcommand{\ourapproach}{\textsc{F2STrans}\xspace}

\newcommand{\eg}{\emph{e.g.,}\xspace}

\definecolor{lightyellow}{RGB}{241,230,209}
\definecolor{blue}{RGB}{218,232,252}

\theoremstyle{plain}

\theoremstyle{definition}

\theoremstyle{remark}

\usepackage[textsize=tiny]{todonotes}

\icmltitlerunning{Function-to-Style Guidance of LLMs for Code Translation}

\begin{document}
	
	\twocolumn[
	\icmltitle{Function-to-Style Guidance of LLMs for Code Translation}
	
	
	
	\icmlsetsymbol{equal}{*}
	\begin{icmlauthorlist}
		\icmlauthor{Longhui Zhang}{hitsz}
		\icmlauthor{Bin Wang}{hitsz}
		\icmlauthor{Jiahao Wang}{hitsz}
		\icmlauthor{Xiaofeng Zhao}{huawei}
		\icmlauthor{Min Zhang}{huawei}
		\icmlauthor{Hao Yang}{huawei} \\
		\icmlauthor{Meishan Zhang}{hitsz}
		\icmlauthor{Yu Li}{zju}
		\icmlauthor{Jing Li\textsuperscript{\faEnvelope}}{hitsz}
            \icmlauthor{Jun Yu}{hitsz}
            \icmlauthor{Min Zhang}{hitsz}
	\end{icmlauthorlist}
	
	\icmlaffiliation{hitsz}{Harbin Institute of Technology, Shenzhen, China.}
	\icmlaffiliation{huawei}{Huawei Translation Services Center, Beijing, China.}
    \icmlaffiliation{zju}{Zhejiang University, Hangzhou, China}
        
	\icmlcorrespondingauthor{Jing Li}{jingli.phd@hotmail.com}
	
	\icmlkeywords{Machine Learning, ICML}
	
	\vskip 0.3in
	]
	
	
	
	\printAffiliationsAndNotice{}  
        
	\begin{abstract}
Large language models (LLMs) have made significant strides in code translation tasks. 
However, ensuring both the correctness and readability of translated code remains a challenge, limiting their effective adoption in real-world software development.
In this work, we propose \ourapproach,  a \textit{function-to-style guiding paradigm} designed to progressively improve the performance of LLMs in code translation. 
Our approach comprises two key stages: (1) \textit{Functional learning}, which optimizes translation correctness using high-quality source-target code pairs mined from online programming platforms, and (2) \textit{Style learning}, which improves translation readability by incorporating both positive and negative style examples.
Additionally, we introduce a novel code translation benchmark that includes up-to-date source code, extensive test cases, and manually annotated ground-truth translations, enabling comprehensive functional and stylistic evaluations.
Experiments on both our new benchmark and existing datasets demonstrate that our approach significantly improves code translation performance. 
Notably, our approach enables Qwen$_{1.5B}$ to outperform prompt-enhanced Qwen$_{32B}$ and GPT-4 on average across 20 diverse code translation scenarios.
\end{abstract}
	\section{Introduction}
Code translation involves converting code from one programming language to another, a task usually required for application porting or software migration~\cite{nguyen2013lexical}. 
Traditionally, this task primarily relied on rule-based methods that required skilled programmers to handle complex cases manually~\cite{zhong2010mining}.
The advent of deep learning has led to the development of various learning-based strategies. 
A prominent approach involves pretraining on monolingual code datasets, followed by fine-tuning with bilingual corpora to improve translation accuracy~\cite{lachaux2020unsupervised, wang2021codet5}. 
Although learning-based techniques have significantly outperformed traditional rule-based methods, their performance remains insufficient for real-world deployment~\cite{yang2024exploring}.

Large language models (LLMs), such as GPT-4~\cite{openai2024gpt4technicalreport}, have revolutionized various coding tasks, including code generation~\cite{liu2024exploring} and program repair~\cite{fan2023automated}, with their remarkable performance. 
This success has spurred growing interest among researchers in leveraging LLMs for code translation~\cite{codetransocean}.
Simple prompt-based learning allows LLMs to translate code effectively, and optimized prompts can further enhance their performance. 
For example, \citet{rag1, rag2} improved translation accuracy using a retrieval-augmented generation (RAG) strategy. 
\citet{yang2024exploring, pan2024lost} utilized compiler feedback to iteratively refine translations.
Powered by  LLMs, current code translation models have reached unprecedented levels of performance, surpassing traditional approaches~\cite{tao2024unraveling}.

\begin{figure}[t]
	\centering
	\includegraphics[width=0.47\textwidth]{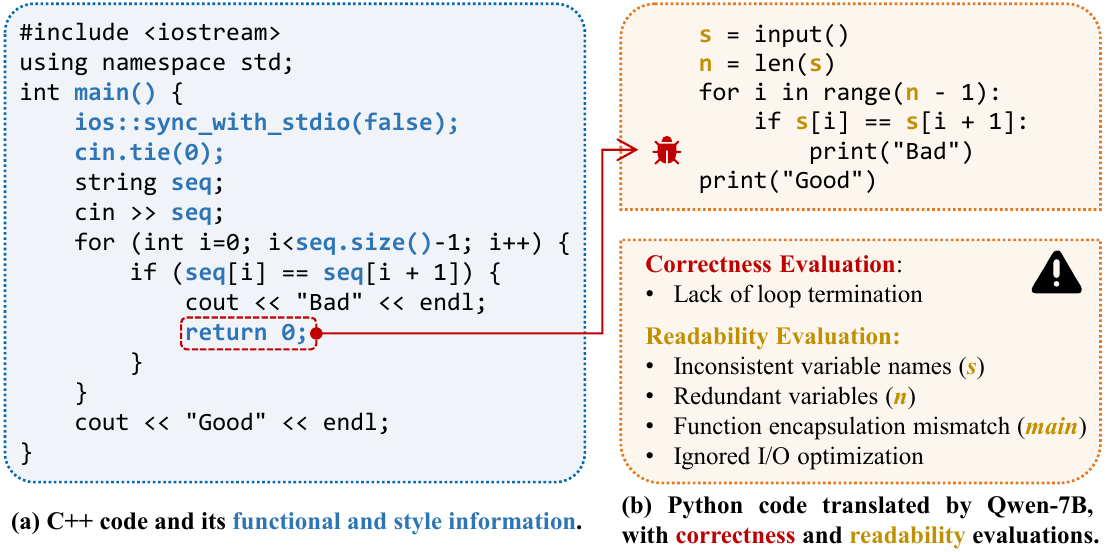}
	\caption{Limitations in the correctness and readability of code translation by LLMs.}
	\label{fig:intro}
\end{figure}

While the effectiveness of LLMs in code translation tasks is widely acknowledged, most models face two critical limitations, as illustrated in Figure~\ref{fig:intro}:
(i) \textbf{Correctness}: For instance, StarCoder$_{3B}$ achieves an average success rate of only 7\% on the traditional code translation benchmark CodeNet~\cite{pan2024lost}.
(ii) \textbf{Readability}: 
Even when translations are functionally correct, they often fail to preserve the source code's style, including code structure and variable naming conventions~\cite{weisz2022better}. 
This lack of readability imposes a substantial burden on developers, as reading poorly structured code often takes longer than writing it from scratch~\cite{martin2009clean}.
Undeniably, more powerful LLMs can produce higher-quality code translations, but they come with notable drawbacks, such as massive model sizes (\eg Qwen$_{32B}$~\cite{qwen2025qwen25technicalreport}) or limited accessibility (\eg GPT-4). 
Relying solely on the inherent capabilities of LLMs to overcome these issues is only a short-term solution.

To address these challenges, we propose \ourapproach, a two-stage guidance framework to improve both correctness and readability: \textit{functional learning} followed by \textit{style learning}. 
Functional learning involves training the model to generate target code that preserves the functionality of the source code.
To achieve this, \ourapproach mines cross-language code pairs from online programming platforms, selects pairs with consistent solutions and functionality, and utilizes these pairs for functional learning.
Style learning ensures the model accurately preserves the stylistic features of the source code in the translated target code, thereby improving readability. 
In this stage, \ourapproach generates target code samples exhibiting both good and poor stylistic quality, enabling the model to recognize and prioritize maintaining stylistic consistency through the style learning.

We conduct extensive experiments to validate our approach. 
To overcome the limitations of existing benchmarks, such as outdated source code, insufficient test cases, and missing ground-truth translations, we construct a new benchmark.
Utilizing this new benchmark, along with the traditional benchmarks, we evaluate \ourapproach across 20 code translation scenarios encompassing five programming languages: C, C++, Go, Java, and Python. 
The results demonstrate that our approach is effective across LLMs of varying types and scales, including StarCoder$_{3B}$ and Qwen$_{0.5-7B}$.
Notably, with our approach, Qwen$_{1.5B}$ surpasses GPT-4 in code translation tasks.

We summarize the key contributions of our work as follows:
\begin{itemize}
\item We propose \ourapproach, a function-to-style guidance framework designed to enhance both the correctness and readability of code translations generated by LLMs.
\item We create a comprehensive benchmark to rigorously evaluate the functional accuracy and stylistic consistency of code translations.
\item Our approach significantly improves the quality of translated code across various types and sizes of LLMs in 20 investigated translation scenarios.
\end{itemize}
	\section{Methodology}
\begin{figure*}[t]
	\centering
	\includegraphics[width=0.95\textwidth]{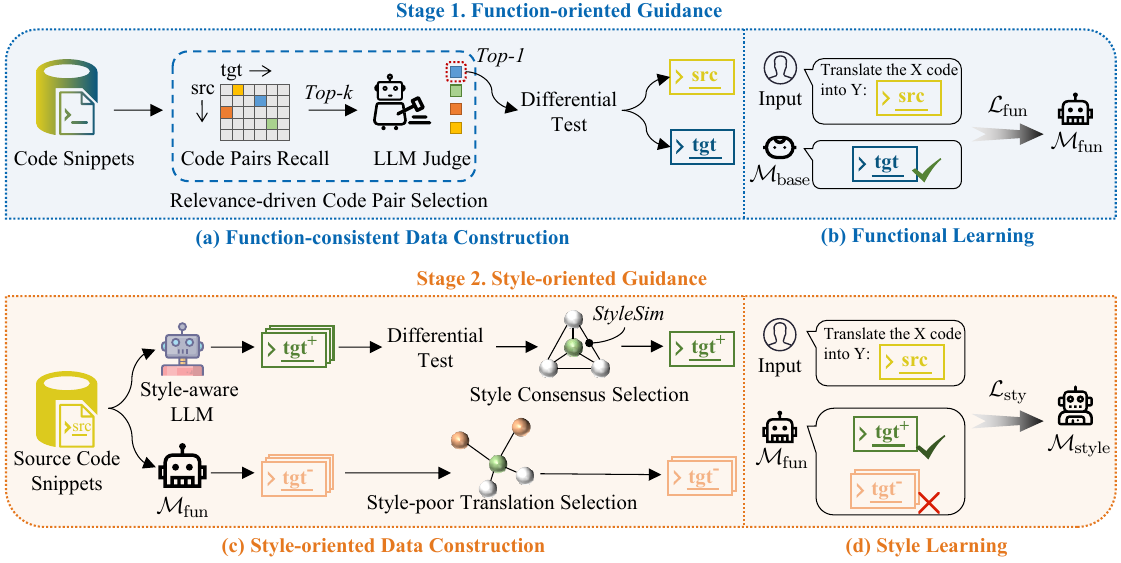}
	\caption{Overview of our \ourapproach. 
Firstly, the base LLM $\mathcal{M}_{\text{base}}$ is transformed into an intermediate model $\mathcal{M}_{\text{fun}}$ after function-oriented guidance. 
Subsequently, $\mathcal{M}_{\text{fun}}$ is refined into the final model $\mathcal{M}_{\text{sty}}$ through style-oriented guidance.}
	\label{fig:main}
	\vspace{-3.0mm}
\end{figure*}

As illustrated in Figure~\ref{fig:main}, \ourapproach employs a two-stage progressive learning paradigm.
First, to ensure functionally consistent code translation, we perform a functional learning using function-consistent code pairs.
Second, to improve the style alignment between source and target code, we propose a novel style-learning mechanism based on positive and negative style translation examples.

\subsection{Function-oriented Guidance}
A fundamental requirement of code translation is functional consistency---ensuring that source and target code produce identical outputs for the same inputs. 
A simple strategy~\cite{xcodeeval} is to fine-tune LLMs using source and target language code that solves the same programming problem on online programming platforms.
Platforms such as Codeforces host numerous programming problems, along with test cases and code solutions submitted by developers worldwide, making this strategy feasible.

However, this strategy can create ambiguity in the model's understanding of the optimal translation, since programming problems typically have multiple solutions.
More critically, solution inconsistencies may lead to divergent output behaviors between technically correct code solutions.
For example, while both Dijkstra's and Floyd-Warshall's algorithms solve the shortest path problem, they may produce different but equally valid results for the same graph.
Obviously, training on code pairs with inconsistent outputs is sub-optimal.
These limitations motivate us to enhance the quality of the data generated by this strategy.

\lparagraph{Function-consistent Data Construction.}
To develop our dataset, we first select highly relevant source-target code pairs from a large number of solutions to the same programming problem, and then retain those exhibiting identical input-output behavior.

$\blacktriangleright$ \emph{Relevance-driven Code Pair Selection.}
To balance effectiveness and efficiency, we employ a two-step method to identify highly relevant code pairs. 
First, we use a lightweight code embedding model \href{https://huggingface.co/jinaai/jina-embeddings-v2-base-code}{Jina}~\cite{jina} to retrieve the top ten similar code pairs based on cosine similarity of embeddings.
Next, we apply a rating scale–based method where an LLM judge assesses the solution consistency of code pairs using fine-grained labels. 
The scoring process is defined as:
\begin{equation}
	\label{eq:recall_rerank}
	\small
	\begin{aligned}
		\mathop{score}\left(src, tgt\right) &= \sum_{k}\mathop{p}\left(k\mid \mathcal{P}_s(src, tgt) \right) \cdot k \\
		\mathop{p}\left(k\mid \mathcal{P}_s(src, tgt) \right) &= \frac{\exp\left(s_{k}\right)}{\sum_{k'}\exp\left(s_{k'}\right)},
	\end{aligned}
\end{equation}
where we use \href{https://huggingface.co/Qwen/Qwen2.5-Coder-7B}{Qwen$_{7B}$} as the LLM judge. 
Here, $src$ and $tgt$ represent source and target code, respectively, while $\mathcal{P}_s(src, tgt)$ is the prompt for the LLM judge, shown in Appendix~\ref{sec:prompt_ours}.  
The variable $k$ is an integer ranges from $1$ to $K$, with higher values signifying greater relevance, and $s_k$ is the log-likelihood score of label $k$ generated by the LLM judge when prompted with $\mathcal{P}_s(src, tgt)$. 
We use fine-grained labels $\{1, \dots, K\}$ instead of binary labels, because code pairs for the same programming problem often exhibit varying degrees of solution similarity. 
Besides, to ensure each code pair receives a distinct score, we aggregate the log-likelihood scores  $s_*$ of all labels to obtain continuous relevance scores.

$\blacktriangleright$ \emph{Differential Testing.}
To ensure functional consistency, we perform differential testing~\cite{mckeeman1998differential} on the most  relevant code pairs. 
This process involves executing identical input on both the source and target code and comparing their outputs for discrepancies. 
Only code pairs whose input and output behavior are exactly the same are retained.

\lparagraph{Functional Learning.}
Instruction Fine-tuning (IFT) involves training LLMs on instruction-output pairs with a next-token prediction objective to enhance the models' ability to follow instructions~\cite{zhang2024instructiontuninglargelanguage}. 
We preliminary improve the code translation performance of base LLM $\mathcal{M}_\text{base}$ by applying IFT on the constructed function-consistent code translation data as follow:
\begin{equation}
	\small
	\mathcal{L}_{\text{fun}}(scr,tgt) = - \sum_i \log p\left({tgt}_i \mid \mathcal{P}\left({src}\right),{tgt}_{<i}\right),
	\label{eq:pretrain_loss}
\end{equation}
where $tgt_{i}$ denotes the $i$-th token of target code $tgt$, $tgt_{<i}$ represents the token sequence preceding the $i$-th token in $tgt$, and $\mathcal{P}(src)$ is a code translation prompt designed to translate the source code $src$.
The trained model is $\mathcal{M}_{\text{fun}}$.

\subsection{Style-oriented Guidance}
Although our functional guidance ensures data quality at the solution and function levels, style inconsistencies inevitably persist in code pairs derived from online programming platforms.
These inconsistencies usually include variations in variable naming, function signatures, code structure, and comments between source and target code.
Training on such style-inconsistent data can limit the model's learning of the code style, thereby diminishing the readability of the translated code.
Therefore, we introduce style-oriented guidance to alleviate this issue.

\lparagraph{Style-oriented Data Construction.}
We construct positive and negative code translation data to help $\mathcal{M}_{\text{fun}}$ discern the desired code style through comparison. 
Positive translations maintain stylistic consistency with the source code, while negative translations do not.

$\blacktriangleright$ \emph{Positive Translation Construction.}
We use a strong LLM \href{https://huggingface.co/Qwen/Qwen2.5-Coder-32B-Instruct}{Qwen$_{32B}$} to generate multiple style-aware translations and then select the optimal translation from these candidates.
Specifically, we first design a style-aware prompt, as detailed in Appendix~\ref{sec:prompt_ours}, which explicitly instructs the LLM to adhere to the stylistic conventions of the source code.
Using this prompt, Qwen$_{32B}$ generates $m$ translations of the source code.
Next, we perform differential testing to filter these translations, retaining only the subset $T^+$ that successfully passes all test cases, thereby ensuring functional correctness.
Finally, we select the target code $tgt^+$ that best preserves the source's style from $T^+$.

A major challenge in this process is how to choose the best translation $tgt^+$ from $T^+$.
While our style-aware prompt contributes to good code style, the occasional style error is still inevitable.
Existing style evaluation methods, such as CSSim~\cite{CSSim}, are limited to code pairs within the same language.
To avoid the use of abnormal code translation, we propose a style consensus selection mechanism, which identifies the optimal translation by selecting the one with the highest stylistic similarity to the other candidate translations.
The selection mechanism can be mathematically represented as follow:
\begin{equation}
	\small
	tgt^{+} = \mathop{\arg\max}\limits_{{tgt}_i \in T^+} \sum_{{tgt}_j\in T^+, i \neq j} \operatorname{StyleSim} \left( {tgt}_i, {tgt}_j \right),
	\label{eq:chosen}
\end{equation}
where $\operatorname{StyleSim}(.)$ is a function that measures the stylistic similarity between two code snippets in the target language. 
We use the CSSim metric~\cite{CSSim} as $\operatorname{StyleSim}$, which quantifies code stylistic similarity based on variable naming, API invocation, and code structure.
A detailed explanation of CSSim can be found in Appendix~\ref{sec:background}.

$\blacktriangleright$ \emph{Negative Translation Collection Construction.}
To ensure that negative translations are representative, we use $\mathcal{M}_{fun}$ as the negative translator. 
Specifically, $\mathcal{M}_{fun}$ first generate multiple candidate negative translations. 
Then, we evaluate their style consistency with the positive translations $tgt^{+}$ using the CSSim metric and retain $n$ negative translations $T^-$ whose CSSim values are less than $\alpha$.

\lparagraph{Style Learning.}
Inspired by contrastive learning~\cite{pmlr-v119-chen20j}, we propose a list-wise loss function to encourage $\mathcal{M}_\text{fun}$ generate style-consistent target code $tgt^+$ while suppressing inconsistent translations $T^-$:
\begin{equation}
	\small
	\begin{aligned}
	\mathcal{L}_{\text{list}}(src, tgt^+, T^-) &=-\log \frac{\exp(\mathcal{S}(src, tgt^+))}{\sum_{tgt \in T} \exp(\mathcal{S}(src, tgt))} \\
	\mathcal{S}(src, tgt)&= \prod_j p\left(tgt_j \mid \mathcal{P}(src),tgt_{<j}\right),
	\end{aligned}
	\label{eq:listloss}
\end{equation}
where $\mathcal{S}(src, tgt)$ denotes the probability that $\mathcal{M}_\text{fun}$ translates the source code $src$ into the target code $tgt$, and $T = T^- \cup \{tgt^+\}$. 
Additionally, we apply IFT on positive translations to emphasize their importance:
\begin{equation}
	\label{eq:styleloss}
	\small
     \mathcal{L}_{\text{sty}} = \beta \cdot \mathcal{L}_{\text{list}}(scr,tgt^+, T^-) + (1-\beta) \cdot \mathcal{L}_{\text{ift}}(scr,tgt^+),
\end{equation}
where $\beta$ is a trade-off hyperparameter ranging between 0 and 1, and $\mathcal{L}_{\text{ift}}$ is the IFT loss, calculated the same way as $\mathcal{L}_{\text{fun}}$ in Eq.~\ref{eq:pretrain_loss}.

	\section{Experiments}

\subsection{Benchmark Construction}
Since pre-training LLMs requires handling vast datasets, traditional benchmarks face the risk of data leakage even after meticulous data cleaning, leading to inaccurate evaluations~\cite{xu2024benchmarkingbenchmarkleakagelarge}.
For example, the latest data for the CodeNet benchmark comes from 2020~\cite{codenet}.
To accurately assess the code translation capabilities of LLMs, a benchmark based on more recent data is essential.

Motivated by this insight, we introduce a new code translation benchmark with three key advantages over traditional benchmarks:
(i) \textbf{Up-to-date source code}: We first collect the most recently released programming problems from Codeforces, and then select up to two code solutions from each problem as source code.
(ii) \textbf{Extensive test cases}: Each source code undergoes extensive manually-annotated test cases that encompass edge conditions and difficult scenarios of programming problems.
(iii) \textbf{Consistent functional and stylistic translations}: We manually translate each source code into the target language to support functional and stylistic evaluation of the translated code.
The detailed statistics presented in Table~\ref{tab:benchmark}.

\begin{table}[t]
	\centering
	\footnotesize 
	\setlength{\tabcolsep}{2pt} 
	\begin{tabular}{lccccc}
		\toprule
		& \bf \#Lang & \bf \#Code & \bf Date & \bf \#Cases & \bf GT \\ 
		\midrule
		CodeNet & $5$ & $200 \times 20$ & Pre-2021 & $1 \times 20$ & \ding{55} \\
		\rowcolor{blue}
		\ourapproach (Ours)    & $5$ & $1000 \times 20$ & Mid-2024 & $50 \times 20$ & \ding{51} \\
		\bottomrule
	\end{tabular}
	\caption{Comparison of CodeNet and our evaluation benchmark.
Both datasets cover 20 translation scenarios across five languages (Lang). 
Our benchmark surpasses CodeNet in terms of a larger and more up-to-date codebase (Code and Date), extensive test cases (Cases), and manually annotated ground-truth translations (GT).
}
\vspace{-3.0mm}
	\label{tab:benchmark}
\end{table}

\begin{table*}[ht]
	\centering
	\belowrulesep=0pt
	\aboverulesep=0pt
	\setlength\tabcolsep{4pt}  
	\renewcommand{\arraystretch}{1.2} 
	\resizebox{\textwidth}{!}{ 
		\begin{tabu}{c|c | cccc | cccc | cccc | cccc | cccc | c}
			\toprule
			
			\multirow{2}{*}{\bf Method} & \multirow{2}{*}{\bf LLM}  & \multicolumn{4}{c|}{\bf Translation C $\rightarrow$ \{\}} & \multicolumn{4}{c|}{\bf Translation C++ $\rightarrow$ \{\}} & \multicolumn{4}{c|}{\bf Translation Go $\rightarrow$ \{\}} & \multicolumn{4}{c|}{\bf Translation Java $\rightarrow$ \{\}} & \multicolumn{4}{c|}{\bf Translation Py $\rightarrow$ \{\}} & \multirow{2}{*}{\bf Avg.} \\
			\cmidrule{3-6} \cmidrule{7-10} \cmidrule{11-14} \cmidrule{15-18} \cmidrule{19-22}
			&  & \bf C++ & \bf Go & \bf Java & \bf Py & \bf C & \bf Go & \bf Java & \bf Py &\bf  C & \bf C++ & \bf Java & \bf Py & \bf C & \bf C++ & \bf Go & \bf Py & \bf C & \bf C++ & \bf Go & \bf Java \\
			\hline \hline
			\multicolumn{23}{c}{\bf (I) CodeNet Benchmark} \\
			\hline \hline
			
			Direct & \multirow{4}{*}{Qwen$_{32B}$} &88.0 &68.9 &76.4 &61.3 &83.5 &69.0 &81.0 &59.5 &69.4 &74.9 &75.4 &58.3 &71.2 &79.3 &64.2 &61.1 &76.9 &81.4 &69.9 &88.5 &72.9 \\
			CoT & &82.9 &72.4 &75.4 &62.8 &80.5 &61.5 &81.0 &65.5 &75.9 &72.4 &77.9 &63.3 &78.3 &80.3 &58.6 &61.6 &73.4 &82.4 &76.4 &88.4 &73.5 \\
			RAG & &87.5 &76.5 &79.0 &70.1 &83.1 &73.1 &81.7 &65.6 &69.0 &75.4 &76.1 &66.0 &77.4 &78.4 &71.9 &66.3 &84.6 &81.5 &77.0 &87.9 &76.4  \\
			Self-debug & &89.1 &74.4 &77.9 &70.1 &86.6 &71.7 &83.9 &67.3 &74.3 &78.4 &79.9 &68.0 &76.6 &82.5 &68.8 &69.4 &81.0 &85.0 &76.0 &90.5 &77.6 \\ \hdashline
			
			Direct &\multirow{4}{*}{GPT-4} &91.0 &69.5 &80.9 &62.6 &83.3 &68.2 &84.0 &64.0 &77.4 &80.4 &77.9 &69.4 &77.8 &83.9 &66.7 &67.3 &76.9 &84.2 &58.1 &82.3 &75.3 \\
			CoT & &88.8 &73.2 &81.1 &65.2 &82.7 &67.0 &83.7 &68.8 &81.3 &80.8 &80.6 &73.5 &86.3 &86.2 &63.5 &67.1 &73.7 &84.5 &62.6 &83.5 &76.7 \\
			RAG & &91.2 &76.2 &85.3 &72.0 &84.6 &70.6 &84.7 &76.2 &78.3 &77.0 &72.6 &77.7 &83.2 &84.1 &72.9 &72.1 &84.4 &86.3 &64.3 &85.1 &78.9 \\
			Self-debug & &\underline{93.5} &72.5 &87.5 &78.5 &86.5 &72.0 &87.5 &74.0 &\textbf{84.5} &\textbf{88.0} &\underline{88.5} &81.0 &\underline{87.0} &\underline{89.5} &70.0 &78.5 &85.5 &87.0 &62.0 &85.0 &81.9 \\ \hdashline
			
			\rowcolor{blue}
			 & Qwen$_{0.5B}$ &91.0 &78.9 &81.9 &73.9 &84.5 &82.0 &84.5 &75.5 &70.9 &73.9 &77.4 &76.4 &74.2 &85.4 &77.3 &75.8 &74.4 &80.9 &73.4 &85.4 &78.9 \\
			\rowcolor{blue} & Qwen$_{1.5B}$ &93.0 &83.4 &89.4 &83.9 &\underline{90.5} &\underline{90.0} &87.5 &80.0 &79.4 &78.4 &82.4 &83.9 &85.9 &86.9 &\underline{86.4} &82.3 &84.4 &87.4 &78.9 &87.4 &85.1 \\
			\rowcolor{blue} & Qwen$_{3B}$  &\textbf{93.6} &86.4 &\underline{90.5} &82.9 &90.0 &89.0 &\textbf{93.5} &\underline{84.0} &79.4 &81.4 &86.4 &83.9 &85.9 &88.9 &85.9 &84.8 &\underline{89.9} &\underline{88.4} &\underline{85.4} &91.0 &\underline{87.1} \\
			\rowcolor{blue} & Qwen$_{7B}$ &91.5 &\bf 92.5 &\bf 93.5 &\bf 86.5 &\bf 92.0 &\bf 92.5 &\underline{93.0} &\bf 88.0 &\underline{82.9} &\underline{86.9} &\bf 89.4 &\bf 87.9 &\bf 88.4 &\bf 89.9 &\bf 88.4 &\bf 87.4 &\bf 93.0 &\bf 89.4 &\bf 87.4 &\bf 94.5 &\bf 89.8 \\
			\rowcolor{blue}
			\multirow{-5}{*}{\shortstack{\textbf{\ourapproach} \\ \textbf{(Ours)}}}  & StarCoder$_{3B}$ &92.0 &\underline{87.9} &89.9 &\underline{84.4} &\textbf{92.0} &89.5 &91.5 &83.5 &80.4 &81.9 &84.9 &\underline{84.9} &86.4 &85.4 &83.8 &\underline{85.4} &86.9 &85.9 &84.9 &\underline{92.5} &86.7 \\ 
			\hline \hline
			\multicolumn{23}{c}{\bf (II) \ourapproach Benchmark (Ours)} \\
			\hline \hline
			
			Direct & \multirow{4}{*}{Qwen$_{32B}$} &81.7 &44.4 &68.5 &48.5 &57.4 &33.6 &56.4 &39.1 &63.2 &69.9 &61.3 &48.7 &71.6 &77.1 &48.1 &63.1 &32.7 &37.1 &32.1 &43.5 &53.9 \\
			CoT & &80.3 &43.7 &68.2 &51.9 &54.4 &34.3 &56.0 &42.0 &66.4 &72.7 &64.9 &53.8 &68.0 &78.5 &44.4 &63.4 &29.9 &40.9 &37.0 &45.5 &54.8 \\
			RAG & &81.3 &57.9 &67.6 &57.2 &56.6 &46.6 &55.3 &41.6 &61.4 &69.3 &60.9 &54.6 &71.1 &77.3 &59.9 &69.2 &34.3 &37.8 &34.1 &44.0 &56.9 \\
			Self-debug & &83.1 &49.0 &69.8 &56.0 &60.4 &35.7 &59.0 &47.2 &68.5 &72.2 &65.8 &57.9 &77.8 &80.1 &51.3 &71.8 &35.3 &39.4 &37.3 &46.0 &58.2 \\ \hdashline
			
			Direct &\multirow{4}{*}{GPT-4} &89.9 &58.1 &81.3 &51.4 &68.9 &52.6 &70.7 &52.6 &70.9 &81.0 &74.1 &68.9 &77.2 &87.7 &60.9 &62.4 &34.4 &44.4 &38.3 &46.7 	&63.6  \\
			CoT & &88.2 &60.2 &80.8 &54.0 &69.2 &50.2 &71.2 &56.5 &75.1 &82.2 &75.7 &72.7 &81.5 &89.2 &57.4 &63.1 &34.2 &42.8 &40.5 &44.5 &64.5 \\
			RAG & &89.7 &61.4 &\underline{87.0} &59.7 &69.2 &56.0 &72.2 &61.8 &71.4 &80.8 &73.4 &75.1 &81.0 &87.0 &64.6 &67.5 &39.5 &47.6 &42.5 &49.3 	&66.8 \\
			Self-debug & &91.3 &61.5 &85.6 &63.2 &71.4 &55.7 &74.1 &61.4 &72.4 &\underline{83.6} &\underline{78.7} &76.7 &\textbf{84.5} &88.7 &67.4 &72.9 &\underline{40.6} &46.8 &42.8 &\bf 51.7 &68.5  \\ \hdashline
			
			\rowcolor{blue}
			& Qwen$_{0.5B}$ &91.0 &71.5 &76.1 &64.8 &66.0 &59.1 &60.6 &53.2 &64.8 &74.8 &67.3 &69.2 &70.3 &84.0 &73.9 &72.8 &25.1 &35.1 &27.6 &34.1 &62.1 \\
			\rowcolor{blue} & Qwen$_{1.5B}$ &93.5 &75.7 &83.6 &71.3 &75.1 &68.2 &71.2 &62.3 &70.7 &79.9 &73.2 &76.6 &79.5 &89.2 &83.7 &82.1 &35.1 &44.8 &35.7 &44.6 &69.8 \\
			\rowcolor{blue} & Qwen$_{3B}$  &\textbf{95.2} &\underline{81.3} &86.7 &\underline{76.1} &\underline{78.2} &\underline{71.8} &\underline{75.4} &\underline{68.9} &75.2 &82.5 &77.4 &\underline{80.7} &81.1 &\textbf{91.1} &86.8 &\underline{84.6} &38.9 &\underline{48.4} &\underline{43.6} &50.4 &\underline{73.7} \\
			\rowcolor{blue} & Qwen$_{7B}$ &\underline{94.6} &\bf 84.0 &\bf 87.6 &\underline{76.1} &\bf 79.5 &\bf 75.8 &\bf 77.6 &\bf 69.4 &\underline{76.4} &\bf 85.1 &\bf 85.1 &\bf 82.0 &\underline{82.7} &\underline{91.0} &\bf 88.2 &\bf 84.7 &\bf 43.2 &\bf 56.5 &\bf 47.5 &\underline{51.3} &\bf 75.9 \\
			\rowcolor{blue} 
			 \multirow{-5}{*}{\shortstack{\textbf{\ourapproach} \\ \textbf{(Ours)}}} & StarCoder$_{3B}$  &94.3 &78.8 &86.5 &\textbf{76.3} &77.8 &71.0 &75.1 &66.7 &\textbf{76.6} &82.0 &77.2 &80.2 &82.0 &90.2 &\underline{87.0} &83.4 &37.9 &45.2 &42.9 &46.5 &72.9  \\ 
			\bottomrule
		\end{tabu}
	}
	\caption{Code translation results of various models on CodeNet and our benchmark. 
	Following standard practice, we adopt CA as the evaluation metric.
	The bold values represent the best results, while the underlined values indicate the second-best.
	}
	\label{tab:main_ex}
\end{table*}

\subsection{Experimental Settings}
\lparagraph{Implementation Details.}
In function-oriented guidance, we set the maximum algorithmic consistency label $K$ in Eq.~\ref{eq:recall_rerank} to 5.
In the style-oriented guidance, we set both the numbers of positive translations $T^+$ and negative translations $T^-$, namely $m$ and $n$, to 10, with the value of $\alpha$ in negative translation collection construction set to 0.8 and the trade-off hyperparameter $\beta$ in Eq.~\ref{eq:styleloss} fixed at 0.6. 
We combine training data from various translation scenarios for mixed training, enabling a single LLM to translate among all the investigated languages.
The LLM-judge prompt $\mathcal{P}_s$ in function-oriented guidance, the style-aware prompt in style-oriented guidance, and the prompt $\mathcal{P}$ of \ourapproach in Eq.~\ref{eq:pretrain_loss} and \ref{eq:listloss} are shown in Appendix~\ref{sec:prompt_ours}.
Appendix~\ref{sec:appendix_implementation} shows more detailed implementation details.

\lparagraph{Baselines and Metric.}
We evaluate our approach against two SOTA LLMs: Qwen$_{32B}$ and GPT-4. 
For each model, we implement four established prompt learning strategies: Direct Prompt Learning~\cite{yang2024exploring}, Chain of Thought (CoT) ~\cite{codetransocean}, RAG~\cite{rag1, rag2}, Self-Debug Prompt Learning ~\cite{yang2024exploring, pan2024lost}.
These strategies have been proposed and validated in related works to enhance model performance in code translation tasks.
Detailed descriptions of each strategy are provided in Appendix~\ref{sec:app_baselines}.

Following standard practice~\cite{lachaux2020unsupervised}, we adopt Computational Accuracy (CA) as our evaluation metric, which measures the proportion of translated code that produce identical execution results to the source code across all inputs.
We allowed each LLM only one translation attempt per source code.
However, for self-debug prompt learning, which inherently requires multiple LLM invocations, we allow an additional iteration for bug fixing.

\subsection{Main Results}
To verify the broad applicability of \ourapproach, our experiments involve various types and sizes of LLMs, including Qwen${_{0.5-7B}}$~\cite{qwen2025qwen25technicalreport} and StarCoder$_{3B}$~\cite{lozhkov2024starcoder}. 
The performance of our model and baselines on CodeNet and our benchmark is presented in Table~\ref{tab:main_ex}, and additional benchmark results in Appendix~\ref{more_results}.

\lparagraph{Evaluation on CodeNet Benchmark.}
From Table~\ref{tab:main_ex}(I), we recognize the immense potential of LLMs in code translation tasks.
For instance, \ourapproach enables Qwen$_{0.5B}$ to surpass RAG-based Qwen$_{32B}$ in average performance across the various translation scenarios examined. 
While \ourapproach lags behind GPT-4 with self-debug prompt learning in some specific cases, such as Go-to-C translation, it is important to consider the increased computational cost associated with the additional reasoning steps required by self-debug prompt learning for bug fixing. 
Therefore, the advantages of \ourapproach remain significant.

\lparagraph{Evaluation on \ourapproach Benchmark.}
As shown in Table~\ref{tab:main_ex}(II), all evaluated models score at least 10 points lower on average in our benchmark compared to CodeNet, highlighting the greater complexity of our proposed benchmark.
An interesting observation is the significantly weaker performance of all models on Python translation within our benchmark. 
This may be attributed to Python's interpreted nature, contrasting with the compiled nature of the other languages. 
This disparity in language types introduces potential challenges in code translation. 
Nonetheless, the overall trend observed on our benchmark is consistent with that of the CodeNet benchmark. 
\ourapproach enables Qwen$_{1.5B}$ to surpass both Qwen$_{32B}$ and GPT-4 in average performance, despite the latter utilizing self-debug prompt learning.
Furthermore, increasing the model size can further enhance performance.

\subsection{Ablation Analysis}
Our ablation analysis quantifies the contribution of each part of \ourapproach to the overall performance improvements.

\begin{figure}[t]
	\centering
	\includegraphics[width=0.48\textwidth]{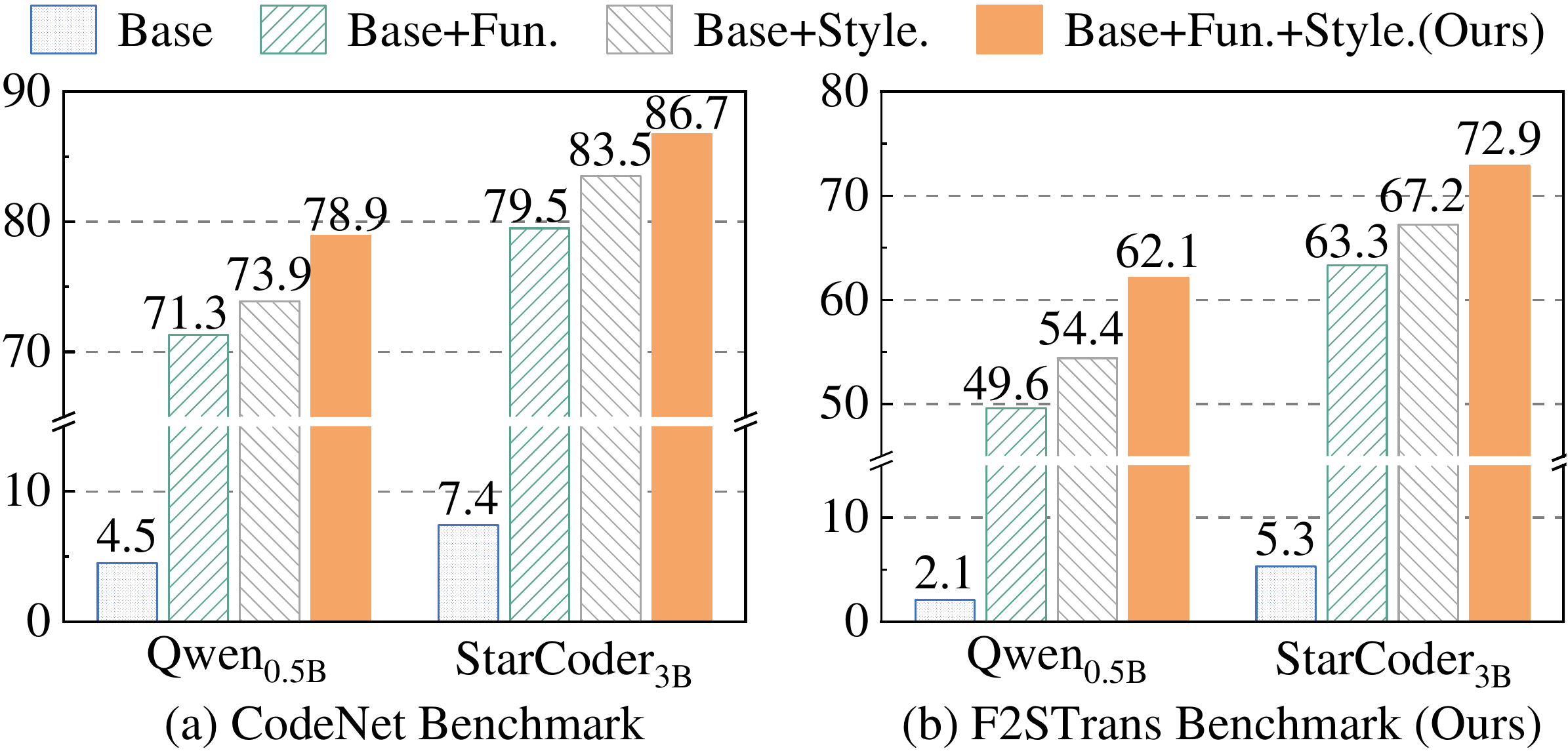}
	\caption{Model performance under different training strategies: base LLMs, function guidance (Fun.), style guidance (Style.), and our function-to-style guidance.}
	\label{fig:ex_fun_style}
	\vspace{-3.0mm}
\end{figure}

\lparagraph{Function-to-Style Guidance.}
Figure~\ref{fig:ex_fun_style} shows the individual contributions of function and style guidance based on Qwen$_{0.5B}$ and StarCoder$_{3B}$.
As shown, both training stages exhibit significant performance gains,
validating their importance to optimal final results.
In particular, the influence of style guidance is highly remarkable.
This is unsurprising, as style guidance involves both positive and negative translations, and the positive translations are derived from Qwen$_{32B}$ rather than online data.
Without these two stages, both Qwen$_{0.5B}$ and StarCoder$_{3B}$ score below 10, underscoring the inherent limitations of naive LLMs in code translation tasks.

\begin{table}[t]
	\centering
	\small
	\belowrulesep=0pt
	\aboverulesep=0pt
	\renewcommand{\arraystretch}{1.1}
	\begin{tabular}{lcccc}
		\toprule
		\multicolumn{1}{c}{\multirow{2}{*}{\bf Method}} & \multicolumn{2}{c}{\bf Qwen$_{0.5B}$} & \multicolumn{2}{c}{\bf StarCoder$_{3B}$} \\
		\cmidrule(r){2-3} \cmidrule(r){4-5}
		&\bf CodeNet &\bf Ours &\bf CodeNet &\bf Ours   \\ \midrule
		\rowcolor{blue}
		\ourapproach (Ours) &\bf 78.9 &\bf 62.1  & \bf 86.7 &\bf 72.9 \\ \midrule		
		\multicolumn{5}{l}{$\bullet $\emph{{ Function-oriented Guidance}}} \\
		\quad w/o RdSel &75.5 &57.5 &83.7 &68.8 \\
		\quad w/o LLM Judge &77.5 	&59.5 	&84.3 	&70.2 \\
		\quad w/o Dif. Test  &78.2 	&60.2 	&85.8 	&71.0 \\ \hdashline
		\multicolumn{5}{l}{$\bullet $\emph{ Style-oriented Guidance}} \\
		\quad w/o StyPro &78.6 &61.5 &86.3 &72.4 \\
		\quad w/o SCS &77.7 &58.6 &84.2 &69.4\\
		\quad w/o SpTS &78.4 &60.3 &86.3 &71.4\\
		\quad w/o $\mathcal{L}_\text{list}$ &74.3 &56.8 &84.1 &68.8\\
		\quad w/o $\mathcal{L}_\text{ift}$ &75.9 &59.0 &85.4 &70.8\\
		\bottomrule
	\end{tabular}
	\caption{Ablation results of our \ourapproach, evaluating the contribution of the following components: 
		 relevance-driven code pair selection (RdSel), LLM judge and differential test (Dif. Test) in function guidance, along with style-aware prompt learning (StyPro.), style consensus selection in Eq.~\ref{eq:chosen} (SCS), style-poor translation selection (SpTS) and the loss function in style guidance.
	}
	\label{tab_ablation}
	\vspace{-3.0mm}
\end{table}

\lparagraph{Function-oriented Guidance.}
The first part of Table~\ref{tab_ablation} examines the impact of three modules in this stage:

$\blacktriangleright$ \emph{Relevance-driven Code Pair Selection.}
To validate the effectiveness of this module, we compared \ourapproach to a variant that randomly selects code pairs, disregarding solution differences. 
The notable performance decline observed with this variant confirms our hypothesis that the solution gaps between code versions are crucial and cannot be ignored. 
Bridging these gaps is essential for achieving optimal model performance.

$\blacktriangleright$ \emph{LLM Judge.}
To assess the impact of LLM judgments, we implemented a variant that directly uses the embedding model Jina to select the top-ranked relevant code pair for training.
The observed performance decline in this model variant indicates that, although Jina is specifically designed for relevance assessment, LLM judgments can effectively refine its retrieval results. 
This further corroborates the strong task generalization capabilities of LLMs.

$\blacktriangleright$ \emph{Differential Testing.}
Although we have strived to maximize the solution relevance of the training data, overlooking differential testing has undeniably adversely affected the training process. 
This phenomenon indicates that assessing the input–output behaviors of code pairs is the most effective approach to verifying their functional consistency.

\begin{figure}[t]
	\centering
	\includegraphics[width=0.45\textwidth]{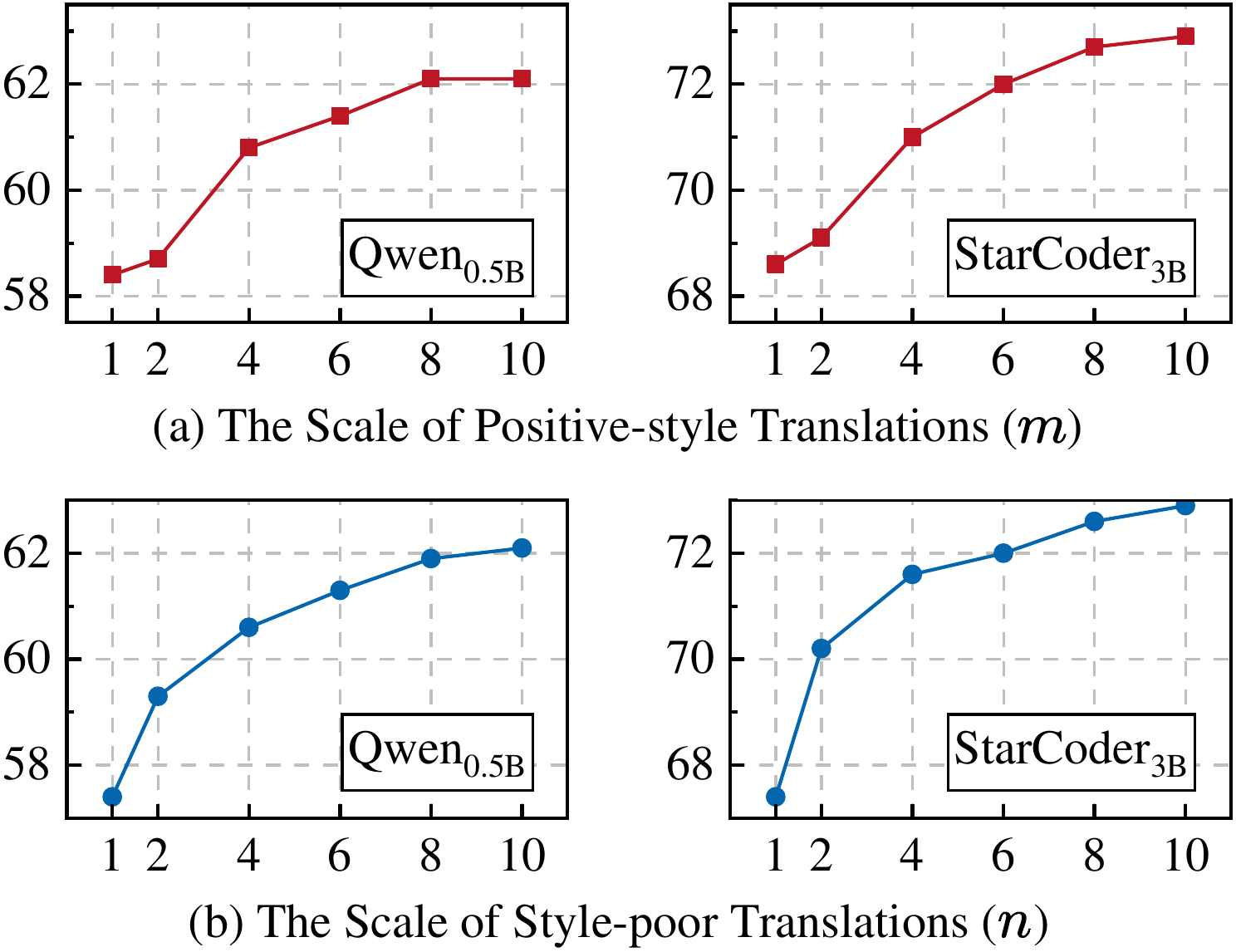}
	\caption{Impact of the number of positive translations generated by Qwen$_{\text{32B}}$ and the number of negative translations used in style learning, based on our benchmark.}	
	\label{fig:ex_mn}
\end{figure}

\lparagraph{Style-oriented Guidance.}
As shown in Table~\ref{tab_ablation} and Figure~\ref{fig:ex_mn}, we evaluate four key modules of style supervision:

$\blacktriangleright$ \emph{Style-aware Prompt Learning.}
We employ style-aware prompts to guide LLMs in generating positive translations that align with the source code style. 
To evaluate the necessity of this configuration, we remove all style-related information from the prompt, thereby allowing the LLMs to generate code without stylistic constraints. 
As shown in Table~\ref{tab_ablation} under ``w/o StyPro'', both Qwen$_{0.5B}$ and StarCoder$_{3B}$ exhibit decreased performance, indicating that well-defined styles in the training data facilitate LLMs' adaptation to code translation tasks.

$\blacktriangleright$ \emph{Style-oriented Data Selection of Positive and Negative Candidate Translations.}
As illustrated in Figure~\ref{fig:main}, we employ a style consensus strategy to select the optimal positive translation from those generated by Qwen$_{\text{32B}}$ and utilize negative translations that significantly differ in style from the positive data.
In the ablation studies presented under the ``w/o SCS'' and ``w/o SpTS'' entries in Table~\ref{tab_ablation}, we omit these two data selection methods and instead randomly select one positive candidate translation and $n$ negative candidate translations for the style learning in Eq.~\ref{eq:styleloss}. 

The results demonstrate that random data is of inferior quality compared to our carefully curated dataset. 
This yields two key insights:
(I) Even functionally correct translations generated by powerful LLMs may contain subtle stylistic inconsistencies. 
(II) Ensuring a clear stylistic distinction between positive and negative translations facilitates more effective style learning.

$\blacktriangleright$ \emph{The Number of Positive and Negative Candidate Translations.}
Figure~\ref{fig:ex_mn} illustrates the influence of these two numbers, $m$ and $n$, based on our benchmark.
We observe that increasing both $m$ and $n$ significantly enhances model performance, with results stabilizing around $m = 8$ and $n = 10$. 
This is expected, as a larger number of positive candidates increases the likelihood of identifying the optimal translation, while more negative translations expose potential deficiencies of $\mathcal{M}_{\text{fun}}$.
Furthermore, the steeper curves demonstrate that scaling these data yields greater benefits for StarCoder$_{3B}$, indicating that larger model sizes achieve higher performance ceilings.

$\blacktriangleright$ \emph{Loss Function.}
Our loss function comprises $\mathcal{L}_{\text{list}}$ and $\mathcal{L}_{\text{ift}}$. 
Ablation studies in Table~\ref{tab_ablation} indicate that $\mathcal{L}_{\text{list}}$ significantly enhances model performance, and the addition of $\mathcal{L}_{\text{ift}}$ further increases these benefits. 
Moreover, $\mathcal{L}_{\text{list}}$ outperforms $\mathcal{L}_{\text{ift}}$ because it not only encourages the generation of positive translations like $\mathcal{L}_{\text{ift}}$ does, but also suppresses the generation of negative translations.

\subsection{Discussion}
In this subsection, we conduct detailed experimental analyses to comprehensively evaluate \ourapproach.

\begin{figure}[t]
	\centering
	\includegraphics[width=0.48\textwidth]{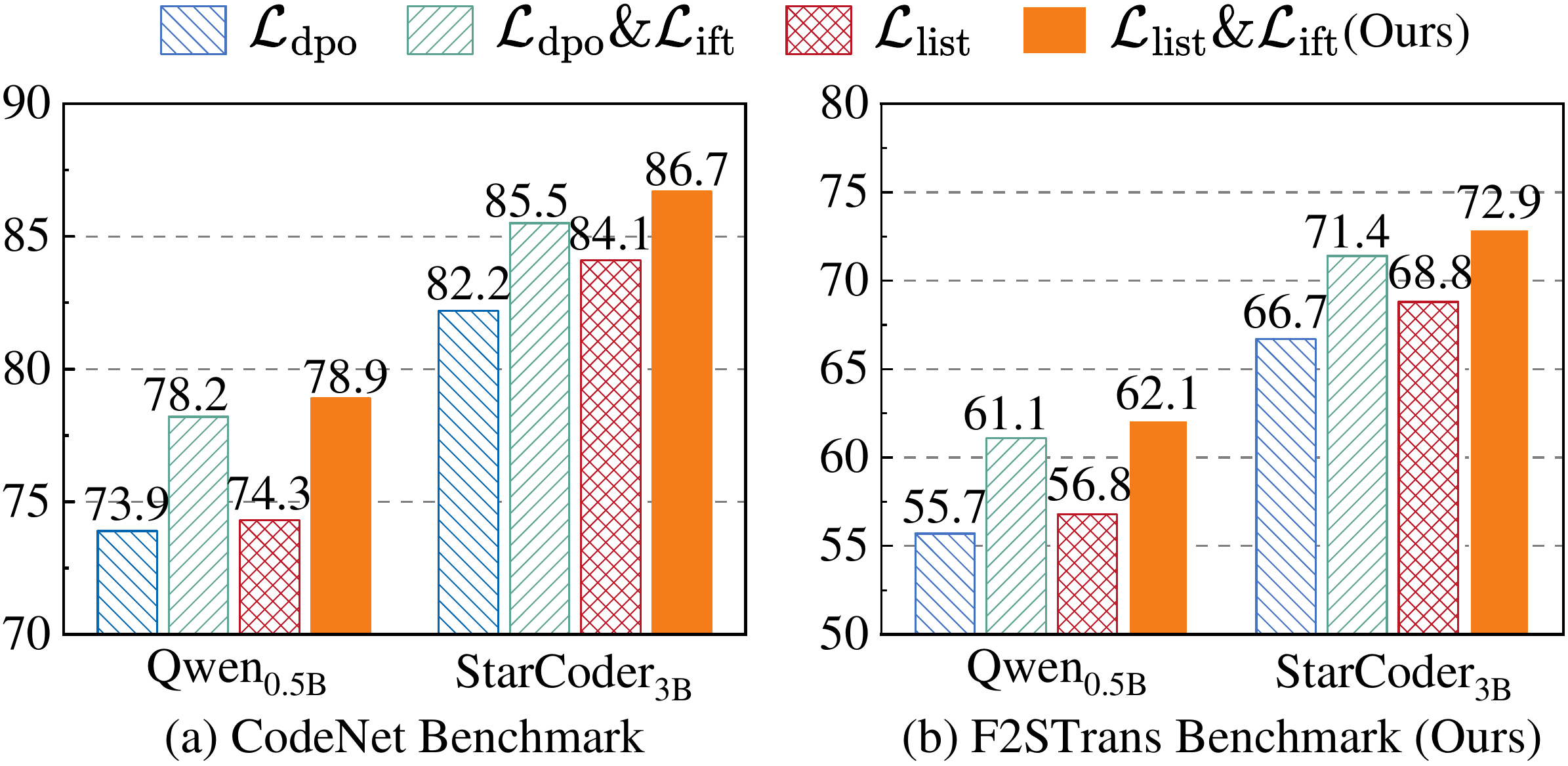}
	\caption{Results of style learning based on various loss functions.}
	\label{fig:ex_dpo}
        \vspace{-3.0mm}
\end{figure}

\lparagraph{A Comparison Between Our List-wise Loss Function  $\mathcal{L}_{\text{list}}$ and Preference Learning Loss Functions.}
Our style learning can be realized using traditional preference learning strategies, such as RLHF~\cite{rlhf} and PPO~\cite{ppo}. 
Here, we compare our list-wise loss function $\mathcal{L}_{\text{list}}$ in Eq.~\ref{eq:listloss} with Direct Preference Optimization (DPO)~\cite{dpo}, a leading preference learning method known for its simple and effective training process. 
Since DPO uses only one positive and one negative data, in our experiment, we align the positive sample of DPO with \ourapproach, while the negative data is randomly selected from the negative data of \ourapproach. 
Additionally, following our Eq~\ref{eq:styleloss}, we also compare the model's performance with the instruction-tuning loss function $\mathcal{L}_{\text{ift}}$ as an auxiliary. 
The results shown in Figure~\ref{fig:ex_dpo} demonstrate that, under both experimental setups, our list-wise loss function $\mathcal{L}_{\text{list}}$ significantly outperforms DPO. 
Furthermore, including $\mathcal{L}_{\text{ift}}$ as an auxiliary also improves DPO performance.

\begin{figure*}[t]
	\centering
	\includegraphics[width=0.9\textwidth]{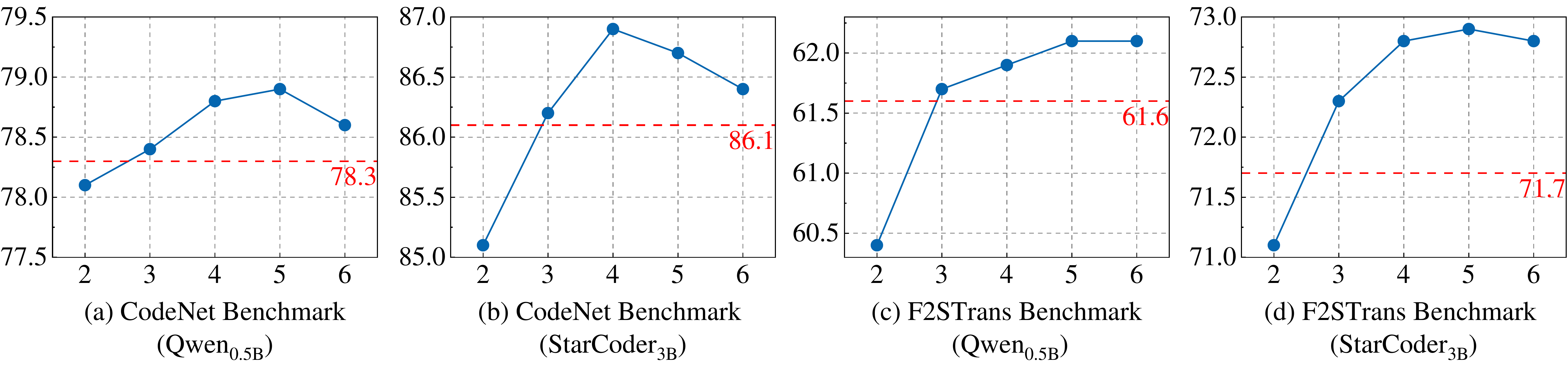}
	\caption{
The impact of the fine-grained label count $K$ (blue curve) and the use of explicit scoring (red dashed line) on the results in the LLM judge. 
In the explicit scoring setting, we set $K$ to 5 and directly use the scores generated by Qwen$_\text{7B}$ as to measure the relevance between source and target code. 
In cases of ties among the highest scores, one is selected randomly.}
	\label{fig:ex_k}
\end{figure*}

\lparagraph{LLM Judge of Function-oriented Guidance.}
In Figure~\ref{fig:ex_k}, we provide an in-depth analysis of LLM judge from two perspectives: the number of fine-grained labels \(K\) and the comparison with explicit scoring.

$\blacktriangleright$ \emph{The Number of Fine-grained Labels $K$.}  
First, compared to binary labels, the advantage of using more fine-grained labels is evident.  
Performance initially improves with increasing label details, peaking around $K=5$, and then declines.
This suggests that while finer labels enable LLMs to make more accurate judgments about data quality, excessive refinement may cause confusion in the model.

$\blacktriangleright$ \emph{The Comparison with Explicit Scoring.}
Unlike the log-likelihood score that integrates all labels as shown in Eq.~\ref{eq:recall_rerank}, here we directly use the labels generated by LLMs, with the number of labels $K$ set to 5.  
It is observed that explicit scoring performs worse than our method, indicating that the log-likelihood score contains more valuable information than a simple LLM-generated response.  
Moreover, even with explicit scoring, the results outperform those achieved with binary labels.  
This further highlights the importance of fine-grained labeling.

\lparagraph{Evaluation from a Style Perspective.}
\begin{table}[t]
	\centering
	\small
	\belowrulesep=0pt
	\aboverulesep=0pt
	\begin{tabular}{lcccc}
		\toprule
		&\bf Dis$_\mathrm{var}$ $\downarrow$ &\bf Dis$_\mathrm{api}$ $\downarrow$ &\bf Dis$_\mathrm{stru}$ $\downarrow$ &\bf CSSim $\uparrow$  \\ \midrule
		\multicolumn{5}{l}{$\bullet $\emph{ Qwen$_{32B}$}} \\
		Direct &17.7 &24.6 &28.6 &76.4 \\
		CoT &18.9 &24.3 &28.1 &76.2 \\
		RAG &16.5 &23.6 &27.2 &77.6 \\
		Self-debug &16.3 &23.4 &27.0 &77.8 \\ \hdashline
		\multicolumn{5}{l}{$\bullet $\emph{ GPT-4}} \\
		Direct &16.8 &23.2 &26.4 &77.9 \\
		CoT &17.3 &22.8 &25.8 &78.0 \\
		RAG &15.4 &21.8 &24.8 &79.3 \\
		Self-debug &15.2 &21.8 &24.5 &79.5 \\ \hdashline
		\rowcolor{blue}
		\multicolumn{5}{l}{$\bullet $\emph{ \ourapproach (Ours)}} \\
		\rowcolor{blue}
		Qwen$_{0.5B}$ &13.8 &20.1 &23.8 &80.7 \\
		\rowcolor{blue}
		Qwen$_{1.5B}$ &12.7 &19.1 &22.1 &82.0 \\
		\rowcolor{blue}
		Qwen$_{3B}$ &{11.9} &18.5 &{20.8} &{82.9} \\
		\rowcolor{blue}
		Qwen$_{7B}$ &\textbf{11.1} &\bf 17.9 &\textbf{19.6} &\bf 83.8 \\
		\rowcolor{blue}
		StarCoder$_{3B}$ &12.0 &{18.3} &21.1 &82.8 \\
		\bottomrule
	\end{tabular}
	\caption{
Style evaluation of generated translations against the ground truth in our benchmark. 
The assessment measures code differences in variable naming $\mathrm{Dis}_{\mathrm{var}}$, API invocation $\mathrm{Dis}_{\mathrm{api}}$, and code structure $\mathrm{Dis}_{\mathrm{stru}}$, with CSSim integrating all three aspects.
Details of these metrics are shown in Appendix~\ref{sec:background}.}
	\label{tab_style} 
    \vspace{-3.0mm}
\end{table}

Using our manually annotated target code within our benchmark as references, we assess the stylistic quality of translations generated by various models, employing the CCSim metric~\cite{CSSim}.
The results presented in Table~\ref{tab_style} demonstrate that \ourapproach significantly improves LLMs' stylistic awareness and generalizes across different scales and types of tested LLMs. 
A compelling piece of evidence is that \ourapproach-improved Qwen$_{0.5B}$ exhibits superior code style compared to self-debug-based GPT-4, despite GPT-4's better functional performance shown in Table~\ref{tab:main_ex}. 
This indicates that functionally correct code can still exhibit stylistic deficiencies. 
Nonetheless, our \ourapproach effectively mitigates this problem.

\begin{figure}[t]
	\centering
	\includegraphics[width=0.48\textwidth]{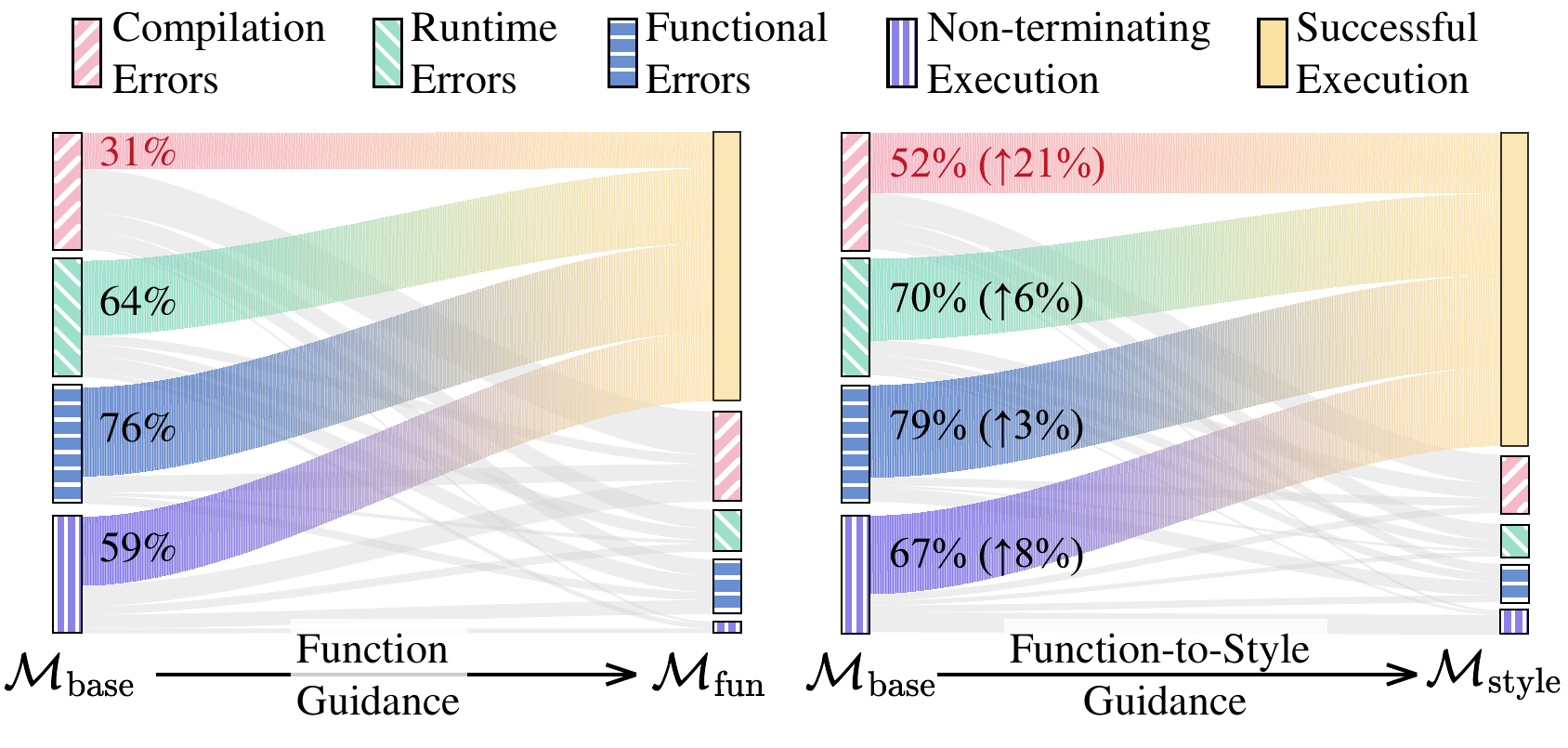}
	\caption{Comparison of error correction in base StarCoder$_{3B}$ translations using function guidance and our function-to-style guidance.
    For the former, 31\% of the compilation errors are successfully corrected, while the latter increases the percentage to 52\%.}
	\label{fig:error}
\end{figure}

\lparagraph{Analysis of \ourapproach in Correcting Errors of Base LLMs.}
Based on our benchmark and the previously established code error classification~\cite{pan2024lost}, we evaluate the effectiveness of function guidance and function-to-style guidance in correcting erroneous translations produced by the naive StarCoder$_{3B}$.
As illustrated in Figure~\ref{fig:error}, function guidance successfully rectifies the majority of the base LLMs' translation errors, while only correcting 31\% of compilation errors.
However, our function-to-style guidance achieves comprehensive improvements across all error types, increasing the correction rate for compilation errors by 21\%. 
This may be because when LLMs diligently adhere to the source code’s style, they can avoid superficial code errors such as undeclared identifiers, which often lead to compilation errors. 
This further underscores the importance of our style guidance.

\lparagraph{Multilingual Modeling Strategies.}
We evaluate four multilingual modeling strategies~\cite{codetransocean} for code translation across the five programming languages used in our work: 
\emph{one2one} (training separate models for each language pair), 
\emph{all2one} (translating code from multiple source languages into one target language using a single model),
\emph{one2all} (translating code from one source language into multiple target languages using a single model),
and \emph{all2all} (a unified model for all language pairs).

\begin{figure}[t]
	\centering
	\includegraphics[width=0.45\textwidth]{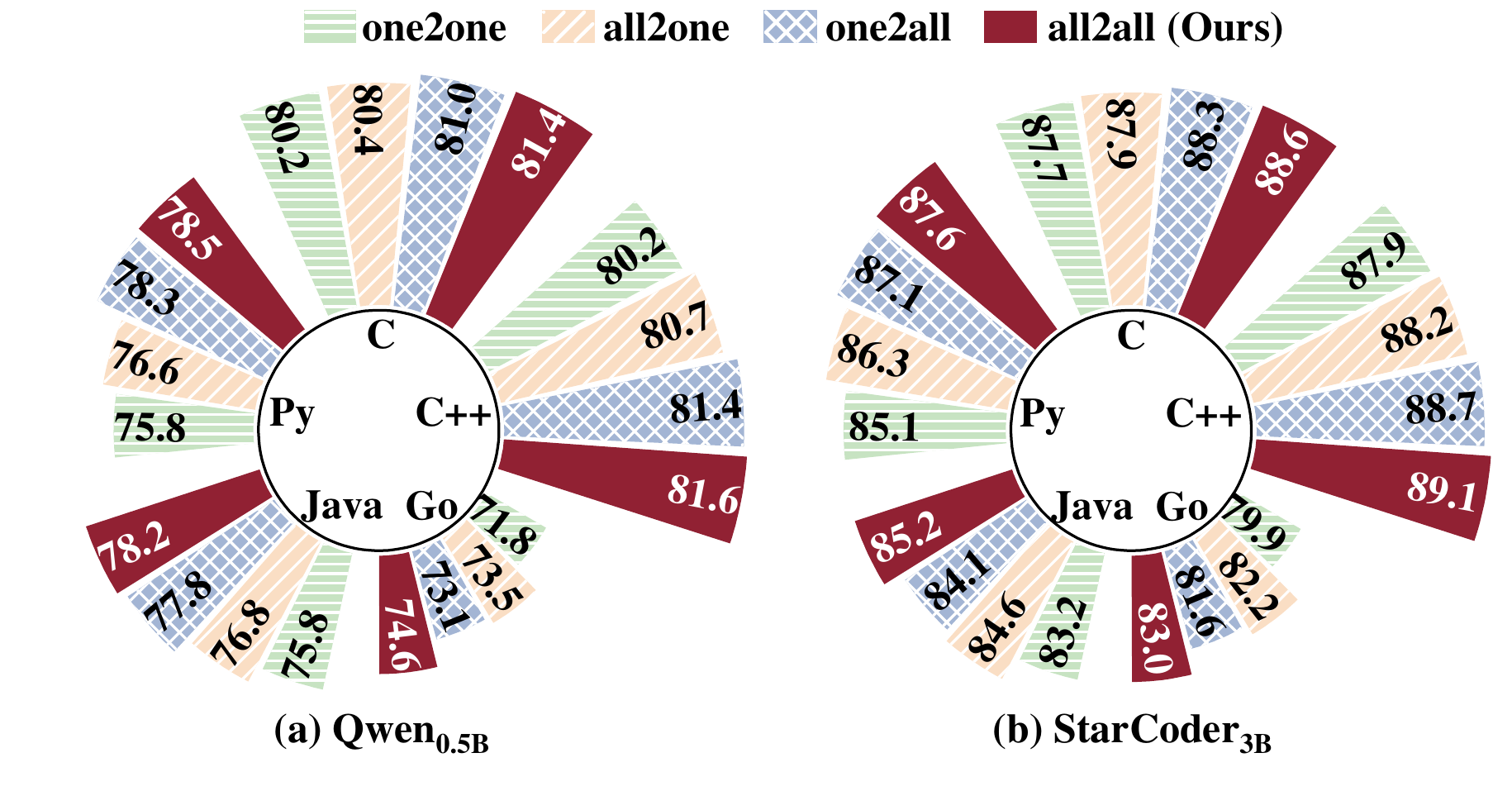}
	\caption{Comparison of various multilingual modeling strategies based on the CodeNet benchmark.}
	\label{fig:ex_mixtraining}
	\vspace{-3.0mm}
\end{figure}

As illustrated in Figure~\ref{fig:ex_mixtraining}, the \emph{one2all}, \emph{all2one}, and \emph{all2all} approaches significantly outperform \emph{one2one}, with \emph{all2all} achieving the highest improvements. 
These results indicate that mixed training across multiple programming languages enhances performance in \ourapproach. 
Notably, translations from Go exhibit the lowest success rates, which can be attributed to Go’s relatively recent emergence in 2009~\cite{donovan2015go}, resulting in limited knowledge within LLMs.
However, the \emph{all2all} strategy provides the most substantial gains when translating from Go, suggesting that it effectively bridges the knowledge gap by enabling LLMs to better understand the relationships between Go and other languages.

        \section{Related Work}
Code translation is essential in software development and maintenance, and has been a subject of extensive research for decades~\cite{mossienko2003automated}. 
Early rule-based and program analysis-based methods, such as CxGo~\cite{c2go}, were costly and required developers to possess a deep understanding of both source and target languages. 
In the era of deep learning, \citet{chen2018tree} minimizes manual intervention by leveraging a learnable attention mechanism to transform the source language’s syntax tree into its counterpart in the target language.
To alleviate the reliance on high-quality parallel code pairs, a series of unsupervised training strategies have been proposed~\cite{DBLP:conf/iclr/Xue0L24, DBLP:conf/iclr/RoziereZCHSL22}.
For example, \citet{lachaux2020unsupervised} employs masked pre-training on large-scale monolingual code data and back-translation strategies to learn the mapping between source and target languages.
Another approach is to use code solutions to identical programming problems in different languages from competitive platforms as weakly supervised parallel data~\cite{ahmad2021avatar, xlcost,codescope,xie2023data}.

Recent advancements in LLM-based code translation have been substantial.
\citet{codetransocean} evaluated ChatGPT’s performance on code translation tasks using standard inference techniques such as direct prompting, few-shot learning, and CoT, demonstrating the effectiveness of LLMs. 
Researchers have since sought to enhance LLMs’ code translation capabilities from multiple angles. 
\citet{macedo2024exploring} explored the impact of output formatting on LLM performance. 
\citet{rag1, rag2} improved translations of low-resource programming languages using a RAG strategy. 
\citet{yang2024exploring, pan2024lost, yin2024rectifier} introduced the self-debugging strategy, where LLM-generated target code is compiled, and any detected bugs are incorporated into subsequent prompts to guide precise fixes. 
\citet{huang2023program, szafraniec2022code, sun2024unicoder} used a unified intermediate representation as a pivot for translating between programming languages, enabling models to capture language-agnostic code semantics effectively. 
Additionally, multi-agent systems~\cite{yuan2024transagent} and human-machine interactive systems~\cite{liu2024hmcodetrans} leveraging LLMs provide developers with more transparent reasoning processes and facilitates the generation of higher-quality translated code.
	\section{Conclusion}
In this study, we introduced a novel feature-to-style training paradigm to effectively adapt LLMs for code translation tasks. 
Initially, we conducted functional learning using high-quality source–target code pairs from online programming platforms, generating functionally correct translations.
Subsequently, we applied style learning based on positive and negative translation data, yielding more readable and stylistically consistent translations.
This two-stage training paradigm significantly improved the performance of LLMs across various scales and types on our newly constructed benchmark as well as traditional code translation benchmarks, even surpassing GPT-4. 
The substantial gains achieved by our approach highlight its effectiveness in advancing code translation capabilities of LLMs. 
        \section*{Impact Statement}
This work introduces a function-to-style guiding paradigm to improve the correctness and readability of code generated by translation models. 
We anticipate that our approach will contribute to more efficient software development and maintenance. 
However, it is crucial to acknowledge the potential for misuse, such as the translation of malicious code. 
Ultimately, adhering to ethical guidelines is essential to ensure the responsible application of this technology.
        \section*{Acknowledgement}
This work was supported in part by National Science Foundation of China (62336008, 62476070, 62125201, U24B20174), Shenzhen Science and Technology Program (JCYJ20241202123503005, GXWD20231128103232001, ZDSYS20230626091203008, KQTD2024072910215406)  and Department of Science and Technology of Guangdong (2024A1515011540).
	
	\bibliography{example_paper}
	\bibliographystyle{icml2025}

	\newpage
	\appendix
	\onecolumn
	\section{Background} \label{sec:background}
\lparagraph{CCSim.}
CCSim, proposed by~\citet{CSSim}, measures stylistic similarity of code by considering the edit distances~\cite{ristad1998learning} of variable naming, API invocation, and code structure.

$\blacktriangleright$ \emph{Calculation of Variable Name Edit Distance --- $\mathrm{Dis}_{\mathrm{var}}$.}
First, extract all variable names, $V_1$ and $V_2$ from the two code respectively. 
Then, compute the edit distance between these two sets of variables as follows:
\begin{equation}
\begin{aligned}
	\mathrm{Dis}_{\mathrm{{V}_1}} & =\frac{1}{||\lambda||_1}\sum_{v_i\in\mathrm{V}_1}\lambda_i\min_{v_j\in\mathrm{V}_2}\mathrm{ED}(v_i,v_j) \\
	\mathrm{Dis}_{\mathrm{V}_2} & =\frac{1}{||\lambda||_1}\sum_{v_i\in\mathrm{V}_2}\lambda_i\min_{v_j\in\mathrm{V}_1}\mathrm{ED}(v_i,v_j) \\
	\mathrm{Dis}_{\mathrm{var}} & =  \frac{\mathrm{Dis}_{\mathrm{V}_1}+\mathrm{Dis}_{\mathrm{V}_2}}{2},
\end{aligned}
\end{equation}
where $\mathrm{ED}$ is the Edit Distance~\cite{ristad1998learning}, and $\lambda_i$ is the normalized inverse document frequency (IDF)~\cite{sparck1972statistical} of a variable naming $v_i$, which is used to decrease the impact of common words.

$\blacktriangleright$ \emph{Calculation of API Invocation Edit Distance --- $\mathrm{Dis}_{\mathrm{api}}$.}
It is calculated similarly to $\mathrm{Dis}_{\mathrm{var}}$, except that variable names are replaced with API names.

$\blacktriangleright$ \emph{Calculation of Code Structure Edit Distance --- $\mathrm{Dis}_{\mathrm{stru}}$.}
The measurement of code structure is based on the Tree Edit Distance (TED)~\cite{paassen2018revisiting} of abstract syntax tree. 
Specifically, it measures the structural difference of two code snippets by determining the fewest insertions, deletions, and replacements needed to transform one tree into the other.

Based on the edit distances of variable names, API invocation, and code structure described above, CCSim measures the stylistic similarity between two code snippets as follows:
\begin{equation}
	\mathrm{CSSim} = 1-\frac{\mathrm{Dis}_{\mathrm{var}}+\mathrm{Dis}_{\mathrm{api}}+\mathrm{Dis}_{\mathrm{stru}}}{3},
\end{equation}
where $\mathrm{CSSim}, \mathrm{Dis}_{\mathrm{var}}, \mathrm{Dis}_{\mathrm{api}}, \mathrm{Dis}_{\mathrm{stru}} \in [0,1]$, and higher CCSim values indicate greater similarity.

\lparagraph{Computational Accuracy.}
We utilize Computational Accuracy (CA) to assess the functional correctness of code translations produced by various models. 
Given all source code, their corresponding target code generated by the models, and the input data $\{(src_1,tgt_1,\text{INPUT}_1),\dots,(src_N,tgt_N,\text{INPUT}_N)\}$, CA is calculated as follows:
\begin{equation}
	\begin{aligned}
\mathrm{CA} & =\frac{\sum_{k=1}^{N} \operatorname{ca}\left(src_{k}, \hat{tgt_{k}}\right)}{N} \\
\operatorname{ca}\left(src_{k}, \hat{tgt_{k}}\right) & =\left\{\begin{array}{ll}
	1, & \text{if } \operatorname{exec}\left(src_{k}, input\right)=\operatorname{exec}\left(tgt_{k}, input\right), \forall input \in \text{INPUT}_k \\
	0, & \text{otherwise}
\end{array}\right.
	\end{aligned}
\end{equation}
where $\operatorname{exec}(.)$ denotes the result of executing the code with a given input.
        \section{More Implementation Details.} \label{sec:appendix_implementation}
In the function-oriented training, we construct approximately 5,000 code pairs for each translation scenario, such as translating from C++ to Python, with a corresponding scale of 10,000 in the style-oriented training.
Throughout both training stages, we maintain consistent hyperparameters, employing 2 epochs and a learning rate of $1 \times 10^{-5}$. 
During inference, we set the temperature of the LLMs to 0.7.
All our experiments are carried out on a machine equipped with eight NVIDIA A800-SXM4-80GB GPUs.

	\section{Baseline Details}
\label{sec:app_baselines}
We adopt Qwen$_{32B}$ and GPT-4 as our baselines, implementing four previously established prompt learning strategies.  
The prompts are shown in Appendix~\ref{sec:prompt_baseline}.
\begin{itemize}[noitemsep,nolistsep]
\item  \textbf{Direct prompt learning (Direct).} 
This straightforward strategy~\cite{yang2024exploring} provides the LLM with a prompt that includes the source language, target language, source code, and a concise task description.
Owing to the powerful instruction-following abilities of LLMs, direct prompt learning has proven highly effective across a broad range of tasks~\cite{du-etal-2024-knowledge,DBLP:conf/pakdd/ShiZ23}. 
\item  \textbf{Chain of thought (CoT).} This strategy, proposed by~\citet{codetransocean}, first encourages the model to thoughtfully consider the translation process and identify potential challenges before undertaking code translation.
This strategy usually achieves better results than direct prompt learning~\cite{lee2024multimodal}.
\item  \textbf{Retrieval-augmented generation (RAG).} Consistent with previous RAG-based code translation approaches~\cite{rag1, rag2}, we retrieve the most similar source-target pair from our style-oriented positive translation dataset using the code embedding model Jina. This retrieved pair provides auxiliary context for the model.
\item  \textbf{Self-debug prompt learning (Self-debug).} 
This method~\cite{yang2024exploring,pan2024lost} employs LLMs to generate initial translations and test cases, verifies the translated code's correctness using these tests, and subsequently corrects any errors based on the test results and compiler error messages.
\end{itemize}

	\section{Additional Results} \label{more_results}
We further evaluate \ourapproach on xCodeEval~\cite{xcodeeval}, as shown in Table~\ref{tab:more_main_ex}. 
We can find that \ourapproach continues to demonstrate a significant advantage. 
Notably, \ourapproach enables Qwen$_{0.5B}$ to outperform both Qwen$_{32B}$ and GPT-4 on average across 20 code translation tasks.  
Qwen$_{1.5B}$ even surpasses the self-debugging GPT-4.
\begin{table*}[t]
	\centering
	\belowrulesep=0pt
	\aboverulesep=0pt
	\setlength\tabcolsep{4pt}  
	\renewcommand{\arraystretch}{1.2} 
	\resizebox{\textwidth}{!}{ 
		\begin{tabu}{cc | cccc | cccc | cccc | cccc | cccc | c}
			\toprule
			
			\multirow{2}{*}{\bf Method} & \multirow{2}{*}{\bf LLM}  & \multicolumn{4}{c|}{\bf Translation C $\rightarrow$ \{\}} & \multicolumn{4}{c|}{\bf Translation C++ $\rightarrow$ \{\}} & \multicolumn{4}{c|}{\bf Translation Go $\rightarrow$ \{\}} & \multicolumn{4}{c|}{\bf Translation Java $\rightarrow$ \{\}} & \multicolumn{4}{c|}{\bf Translation Py $\rightarrow$ \{\}} & \multirow{2}{*}{\bf Avg.} \\
			\cmidrule{3-6} \cmidrule{7-10} \cmidrule{11-14} \cmidrule{15-18} \cmidrule{19-22}
			&  & \bf C++ & \bf Go & \bf Java & \bf Py & \bf C & \bf Go & \bf Java & \bf Py & \bf  C & \bf C++ & \bf Java & \bf Py & \bf C & \bf C++ & \bf Go & \bf Py & \bf C & \bf C++ & \bf Go & \bf Java \\
			\midrule
			Direct & \multirow{4}{*}{Qwen$_{32B}$} &90.3 &57.1 &81.6 &62.5 &73.6 &54.6 &77.0 &56.8 &79.2 &83.1 &80.4 &61.7 &78.1 &75.3 &49.9 &68.6 &70.0 &71.0 &61.5 &78.5 &70.5 \\
			CoT & &85.9 &61.9 &76.6 &63.1 &74.5 &51.9 &76.3 &62.2 &74.7 &83.4 &81.2 &66.3 &75.7 &77.0 &45.0 &70.7 &68.4 &75.6 &65.3 &82.6 &70.9 \\
			RAG & &87.3 &69.5 &78.7 &67.9 &77.7 &65.1 &74.3 &58.2 &83.5 &82.2 &78.5 &68.1 &78.0 &77.1 &60.8 &74.5 &74.5 &76.1 &69.0 &83.0 &74.2 \\
			Self-debug & &91.2 &64.9 &83.2 &67.4 &82.5 &58.6 &78.1 &64.4 &82.2 &85.1 &82.6 &68.2 &79.4 &82.5 &55.8 &77.3 &73.6 &79.7 &69.8 &84.3 &75.5 \\ \hdashline
			
			Direct &\multirow{4}{*}{GPT-4} &95.5 &64.5 &90.2 &65.6 &87.9 &63.5 &85.5 &66.6 &87.5 &91.4 &88.1 &77.4 &84.1 &82.2 &57.4 &71.9 &71.5 &76.0 &58.9 &76.6 &77.1 \\
			CoT & &92.6 &70.7 &86.5 &65.3 &83.7 &62.8 &85.7 &71.0 &85.3 &92.6 &88.6 &80.6 &84.2 &85.4 &54.5 &72.9 &70.6 &77.7 &64.8 &83.8 &78.0 \\
			RAG & &93.5 &72.1 &91.3 &70.1 &84.4 &67.9 &83.9 &74.1 &85.6 &91.5 &85.9 &84.3 &85.1 &84.9 &64.2 &76.2 &77.1 &82.9 &66.8 &84.7 &80.3 \\
			Self-debug & &96.9 &70.2 &90.3 &75.0 &89.2 &68.1 &87.1 &75.1 &\underline{88.9} &\underline{93.4} &89.1 &84.7 &85.6 &\underline{90.2} &64.6 &82.3 &78.3 &81.3 &65.1 &85.0 &82.0 \\ \hdashline
			
			\rowcolor{blue}
			& Qwen$_{0.5B}$ &96.2 &76.0 &87.5 &77.3 &87.0 &73.6 &80.3 &72.1 &80.5 &84.7 &84.2 &81.4 &78.9 &81.2 &68.9 &80.7 &64.9 &70.0 &60.6 &71.9 &77.9 \\
			\rowcolor{blue} & Qwen$_{1.5B}$ &97.3 &83.8 &91.6 &82.7 &90.2 &82.8 &87.1 &78.9 &87.2 &90.3 &90.1 &87.6 &85.6 &86.0 &79.0 &90.3 &76.3 &80.0 &73.7 &85.4 &85.3 \\
			\rowcolor{blue} & Qwen$_{3B}$  &97.3 &\underline{86.6} &\underline{93.9} &83.7 &\underline{91.9} &\underline{85.2} &\underline{89.6} &81.2 &88.2 &92.1 &\underline{91.4} &\underline{89.5} &\underline{86.8} &89.4 &\underline{81.6} &\underline{90.8} &\underline{79.2} &\underline{84.3} &\underline{77.4} &\underline{88.0} &\underline{87.4} \\
			\rowcolor{blue} & Qwen$_{7B}$  &\bf 97.7 &\bf 88.4 &\bf 94.6 &\bf 84.9 &\bf 93.0 &\bf 86.8 &\bf 89.9 &\bf 85.4 &\bf 89.5 &\bf 93.6 &\bf 93.2 &\bf 91.0 &\bf 89.8 &\bf 90.4 &\bf 83.7 &\bf 93.5 &\bf 81.2 &\bf 85.8 &\bf 80.1 &\bf 89.3 &\bf 89.1 \\
			\rowcolor{blue}
			\multirow{-5}{*}{\shortstack{\textbf{\ourapproach} \\ \textbf{(Ours)}}} & StarCoder$_{3B}$ &\underline{97.5} &86.1 &92.4 &\underline{84.8} &91.2 &84.9 &88.9 &\underline{82.1} &88.8 &91.6 &90.9 &88.1 &86.0 &89.1 &80.6 &\underline{90.8} &78.0 &83.0 &76.8 &85.9 &86.9 \\
			\bottomrule
		\end{tabu}
	}
	\caption{Code translation results of various models on XcodeEval benchmarks.}
	\label{tab:more_main_ex}
	\vspace{-3.0mm}
\end{table*}

	\section{Additional Discussion} \label{more_discussion}

\begin{figure}[t]
	\centering
	\includegraphics[width=0.9\textwidth]{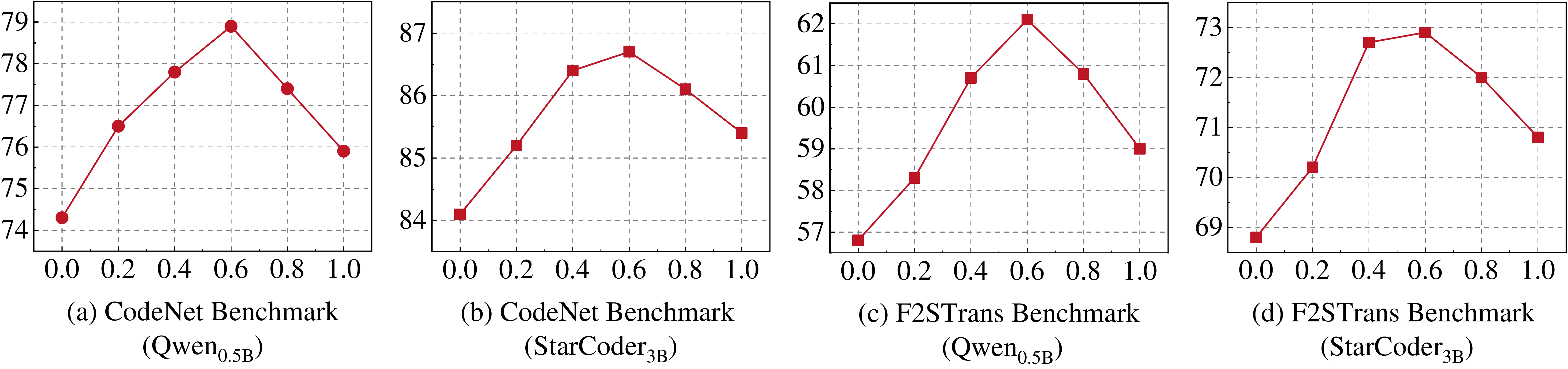}
	\caption{The impact of trade-off hyper-parameters $\beta$ in the loss function Eq.~\ref{eq:styleloss} on the results.}
	\label{fig:ex_beta}
\end{figure}

\lparagraph{The $\beta$ in the Loss Function Eq.~\ref{eq:styleloss} of Style Learning.}
Figure~\ref{fig:ex_beta} illustrates the performance of the model with different trade-off hyperparameters $\beta$. 
It is evident that, during the style learning, as the weight of the instruction fine-tuning loss $\mathcal{L}_{\text{list}}$ increases, the model's performance improves. 
The best performance is achieved when $\beta \approx 0.6$, after which the performance gradually declines. 
Furthermore, when $\beta = 1.0$ (i.e., using only $\mathcal{L}_{\text{list}}$), the model consistently outperforms the case when $\beta = 0.0$ (i.e., using only $\mathcal{L}_{\text{ift}}$). 
This is because $\mathcal{L}_{\text{list}}$ takes both positive and negative data into account, whereas $\mathcal{L}_{\text{ift}}$ considers only the positive data.

	\section{Prompt Settings}

\subsection{Prompts Used by \ourapproach} \label{sec:prompt_ours}

\begin{tcolorbox}[
	colback=myblue!5!white,
	colframe=myblue!75!black,
	arc=1mm, 
	auto outer arc,
	title={LLM-judge Prompt.},
	breakable
	]\small
	
	You are given a source code in \textcolor{red}{\{SOURCE\_LANG\}} and a translated code in \textcolor{red}{\{TARGET\_LANG\}}. Please evaluate the translation by scoring it on a scale from 1 to 5, where:
	
	1: The translated code has significant differences in logic, structure, or implementation compared to the source code, and would likely not work as intended.
	
	2: The translated code works, but there are noticeable differences in logic, style, or structure that deviate from the original solution.
	
	3: The translated code is mostly similar to the source code but has minor differences or optimizations that do not impact overall functionality.
	
	4: The translated code is very close to the original code, with minor, non-critical deviations in style or structure.
	
	5: The translated code is highly consistent with the source code, both in terms of logic and structure, and works as intended.

	\#\#\# \textcolor{red}{\{SOURCE\_LANG\}} Code:
	
	\textcolor{red}{\{SOURCE\_CODE\}}

	\#\#\# \textcolor{red}{\{TARGET\_LANG\}} Code:
	
	\textcolor{red}{\{TARGET\_CODE\}}
	
	\#\#\# Score:
\end{tcolorbox}

\begin{tcolorbox}[
	colback=myblue!5!white,
	colframe=myblue!75!black,
	arc=1mm, 
	auto outer arc,
	title={Style-aware Prompt.},
	breakable
	]\small
	
Translate the following \textcolor{red}{\{SOURCE\_LANG\}} code to \textcolor{red}{\{TARGET\_LANG\}} while preserving the source code style, including variable names, function names, and code structure. Adhere to the following guidelines:

1. Variable and Function Names:

- Maintain the same variable and function names as in the source code.

- If necessary due to language constraints, adjust names minimally while keeping them similar to the original.

2. Code Structure:

- Preserve the overall structure and logic flow of the source code.

- Maintain the same control structures (e.g., loops, conditionals) and their nesting levels.

3. Libraries and APIs:

- Replace source language libraries and functions with equivalent \textcolor{red}{\{TARGET\_LANG\}} libraries and functions.

- Keep the variable and parameter names the same as in the source code where possible.

4. Comments:

- Retain any comments present in the source code.

- Translate comments to \textcolor{red}{\{TARGET\_LANG\}} if applicable, maintaining their position and style.

5. Code Formatting:

- Maintain a similar code formatting style to the source code, including indentation, spacing, and line breaks, as much as possible within the conventions of \textcolor{red}{\{TARGET\_LANG\}}.

Print only the translated \textcolor{red}{\{TARGET\_LANG\}} code and end with the comment ``End of Code''.

\#\#\# Source code:

\textcolor{red}{\{SOURCE\_CODE\}}

\end{tcolorbox}

\begin{tcolorbox}[
	colback=myblue!5!white,
	colframe=myblue!75!black,
	arc=1mm, 
	auto outer arc,
	title={Prompts $\mathcal{P}$ in Eq.~\ref{eq:pretrain_loss} and \ref{eq:listloss} During the Training and Inference of \ourapproach.},
	breakable
	]\small

Translate the \textcolor{red}{\{SOURCE\_LANG\}} code to \textcolor{red}{\{TARGET\_LANG\}} code.

\#\#\# \textcolor{red}{\{SOURCE\_LANG\}} Code:

\textcolor{red}{\{SOURCE\_CODE\}}

\#\#\# \textcolor{red}{\{TARGET\_LANG\}} Code:
	
\end{tcolorbox}

\subsection{Prompts Used by Baselines}\label{sec:prompt_baseline}
\begin{tcolorbox}[
	colback=myblue!5!white,
	colframe=myblue!75!black,
	arc=1mm, 
	auto outer arc,
	title={Direct Prompt Learning.},
	breakable
	]\small
	
Translate the following \textcolor{red}{\{SOURCE\_LANG\}} code to \textcolor{red}{\{TARGET\_LANG\}}. Print only the \textcolor{red}{\{TARGET\_LANG\}} code and end with the comment ``End of Code''.

\#\#\# Source code:

\textcolor{red}{\{SOURCE\_CODE\}}
	
\end{tcolorbox}

\begin{tcolorbox}[
	colback=myblue!5!white,
	colframe=myblue!75!black,
	arc=1mm, 
	auto outer arc,
	title={Chain-of-thought Prompt Learning.},
	breakable
	]\small

First, understand the functionality of the following \textcolor{red}{\{SOURCE\_LANG\}} code and predict the execution output. Then, translate the \textcolor{red}{\{SOURCE\_LANG\}} code into \textcolor{red}{\{TARGET\_LANG\}} while maintaining the same functionality, ensuring that the translated code can be successfully executed.

\#\#\# Source code:

\textcolor{red}{\{SOURCE\_CODE\}}
	
\end{tcolorbox}

\begin{tcolorbox}[
	colback=myblue!5!white,
	colframe=myblue!75!black,
	arc=1mm, 
	auto outer arc,
	title={Retrieval Augmented Generation Prompt Learning.},
	breakable
	]\small
	
Translate the following \textcolor{red}{\{SOURCE\_LANG\}} code to \textcolor{red}{\{TARGET\_LANG\}}. Print only the \textcolor{red}{\{TARGET\_LANG\}} code and end with the comment ``End of Code''.

\#\#\# Source code:

\textcolor{red}{\{SOURCE\_CODE\_EXAMPLE\}}

\#\#\# Target code:

\textcolor{red}{\{TARGET\_CODE\_EXAMPLE\}}

\#\#\# Source code:

\textcolor{red}{\{SOURCE\_CODE\}}

\#\#\# Target code:
	
\end{tcolorbox}

\begin{tcolorbox}[
	colback=myblue!5!white,
	colframe=myblue!75!black,
	arc=1mm, 
	auto outer arc,
	title={Self-debug Prompt Learning When Effect is COMPILE ERROR or RUNTIME ERROR.},
	breakable
	]\small
	
You were asked to translate the following \textcolor{red}{\{SOURCE\_LANG\}} code to \textcolor{red}{\{TARGET\_LANG\}}:

\textcolor{red}{\{SOURCE\_CODE\}}

Your response was the following \textcolor{red}{\{TARGET\_LANG\}} code:

\textcolor{red}{\{TRANSLATED\_CODE\}}

Executing your generated code gives the following error because it is syntactically incorrect:

\textcolor{red}{\{STDERR\}}

Can you re-generate your response and translate the above \textcolor{red}{\{SOURCE\_LANG\}} code to \textcolor{red}{\{TARGET\_LANG\}}. Print only the \textcolor{red}{\{TARGET\_LANG\}} code and do not add any other natural language description in your output. Make sure your generated code is syntactically correct.
\end{tcolorbox}

\begin{tcolorbox}[
	colback=myblue!5!white,
	colframe=myblue!75!black,
	arc=1mm, 
	auto outer arc,
	title={Self-debug Prompt Learning When Effect is INCORRECT OUTPUT.},
	breakable
	]\small
	
You were asked to translate the following \textcolor{red}{\{SOURCE\_LANG\}} code to \textcolor{red}{\{TARGET\_LANG\}}:

\textcolor{red}{\{SOURCE\_CODE\}}

Your response was the following \textcolor{red}{\{TARGET\_LANG\}} code:

\textcolor{red}{\{TRANSLATED\_CODE\}}

Executing your generated code gives the following output:

\textcolor{red}{\{GENERATED\_OUTPUT\}}

instead of the following expected output:

\textcolor{red}{\{EXPECTED\_OUTPUT\}}

Can you re-generate your response and translate the above \textcolor{red}{\{SOURCE\_LANG\}} code to \textcolor{red}{\{TARGET\_LANG\}}. Print only the \textcolor{red}{\{TARGET\_LANG\}} code and do not add any other natural language description in your output. Make sure your generated code is syntactically correct. Your generated \textcolor{red}{\{TARGET\_LANG\}} code should take the following input and generate the expected output:

Input:

\textcolor{red}{\{TEST\_INPUT\}}

Expected Output:

\textcolor{red}{\{EXPECTED\_OUTPUT\}}
\end{tcolorbox}

\end{document}